\documentclass[lettersize,journal]{IEEEtran}
\usepackage{amsmath,amsfonts}
\usepackage{algorithmic}
\usepackage{algorithm}
\usepackage{array}
\usepackage[caption=false,font=normalsize,labelfont=sf,textfont=sf]{subfig}
\usepackage{textcomp}
\usepackage{stfloats}
\usepackage{url}
\usepackage{verbatim}
\usepackage{graphicx}
\usepackage{cite}
\usepackage{orcidlink} 
\usepackage{marvosym}
\newcommand{\ie}{\textit{i.e.}}
\newcommand{\eg}{\textit{e.g.}}
\newcommand{\vs}{\textit{vs.}}
\usepackage{newfloat}
\usepackage{listings}

\usepackage{multirow}
\usepackage{colortbl}
\usepackage{pifont}
\usepackage{tikz}
\usepackage{amssymb}
\usepackage{amsmath}
\usepackage{booktabs}

\newcommand*\circled[1]{\tikz[baseline=(char.base)]{
            \node[shape=circle, fill=white, text=black, draw, inner sep=1pt] (char) {#1};}}

\DeclareCaptionStyle{ruled}{labelfont=normalfont,labelsep=colon,strut=off} 
\floatstyle{ruled}
\newfloat{listing}{tb}{lst}{}
\floatname{listing}{Listing}
\begin{document}

\title{Learning Geometric Invariance for Gait Recognition}

\author{
Zengbin Wang$^*$$^{\orcidlink{0000-0002-9319-905X}}$, 
Junjie Li$^*$$^{\orcidlink{0000-0002-2906-8514}}$, 
Saihui Hou$^{\dag}$$^{\orcidlink{0000-0003-4689-2860}}$,~\IEEEmembership{Member,~IEEE,} 
Xu Liu$^{\orcidlink{0000-0002-0401-1343}}$, 
Chunshui Cao$^{\orcidlink{0000−0001−6634−1682}}$, \\
Yongzhen Huang$^{\orcidlink{0000-0003-4389-9805}}$,~\IEEEmembership{Senior Member,~IEEE,} 
Muyi Sun$^{\orcidlink{0000-0001-9506-7643}}$,  
Siye Wang$^{\orcidlink{0000-0002-7262-3077}}$, 
Man Zhang$^{\ddag}$$^{\orcidlink{0000-0003-3043-2122}}$~\IEEEmembership{Member,~IEEE}
\thanks{This work is jointly supported by the National Natural Science Foundation of China (62276031, 62276025, 62206022, 62476027), the National Key Research and Development Program of China (2023YFF0904700), the Doctoral Innovation Fund of Beijing University of Posts and Telecommunications (CX20242083), and the Fundamental Research Funds for the Central Universities (2253200026).}
\thanks{Zengbin Wang, Junjie Li, Muyi Sun, and Siye Wang are with School of Artificial Intelligence, Beijing University of Posts and Telecommunications, Beijing 100876, China. 
(Email: \{wzb1, hnljj, muyi.sun, wsy\}@bupt.edu.cn).}
\thanks{Man Zhang are with School of Artificial Intelligence, Beijing University of Posts and Telecommunications, and School of Computer and Information Science, Qinghai Institute of Technology. (Email: zhangman@bupt.edu.cn).}
\thanks{Saihui Hou and Yongzhen Huang are with the School of Artificial Intelligence, Beijing Normal University, Beijing 100875, China.(Email: \{housaihui, huangyongzhen\}@bnu.edu.cn).}
\thanks{Xu Liu, Chunshui Cao, and Yongzhen Huang are with Watrix Technology Limited Co. Ltd, Beijing 100088, China. (Email: \{xu.liu, chunshui.cao\}@watrix.ai).}
\thanks{$*$ Equal contributions.}
\thanks{{\dag} Project lead.}
\thanks{{\ddag} Corresponding author.}}

\markboth{Preprint}%
{Shell \MakeLowercase{\textit{et al.}}: A Sample Article Using IEEEtran.cls for IEEE Journals}


\maketitle

\begin{abstract}
    The goal of gait recognition is to extract identity-invariant features of an individual under various gait conditions, \eg, cross-view and cross-clothing.
    Most gait models strive to implicitly learn the common traits across different gait conditions in a data-driven manner to pull different gait conditions closer for recognition.
    However, relatively few studies have explicitly explored the inherent relations between different gait conditions.
    For this purpose, we attempt to establish connections among different gait conditions and propose a new perspective to achieve gait recognition: 
    variations in different gait conditions can be approximately viewed as a combination of geometric transformations.
    In this case, all we need is to \textbf{\emph{determine the types of geometric transformations and achieve geometric invariance, then identity invariance naturally follows.}}
    As an initial attempt, we explore three common geometric transformations (\ie, Reflect, Rotate, and Scale) and design a $\mathcal{R}$eflect-$\mathcal{R}$otate-$\mathcal{S}$cale invariance learning framework, named \textbf{${\mathcal{RRS}}$-Gait}.
    Specifically, it first flexibly adjusts the convolution kernel based on the specific geometric transformations to achieve approximate feature equivariance. Then these three equivariant-aware features are respectively fed into a global pooling operation for final invariance-aware learning.
    Extensive experiments on four popular gait datasets (Gait3D, GREW, CCPG, SUSTech1K) show superior performance across various gait conditions.  
\end{abstract}


\section{Introduction} \label{sec:introduction}
    \IEEEPARstart{G}ait recognition, aiming to capture unique walking patterns of an individual, has become a promising biometric identification technology. 
    Its primary advantage lies in the ability to perform recognition at a long distance without requiring cooperation from the individual~\cite{shen2022comprehensiveSurvey,rani2023gaitsurvey-Rani}.
    Recent progress~\cite{fan2025opengait,ma2024VPGait,Huang2024GaitMoE} has demonstrated remarkable performance under various challenging conditions, such as cross-view and cross-clothing, highlighting its practical significance and potential for real-world applications.

    \begin{figure}
        \centering
        \includegraphics[width=1.0\linewidth]{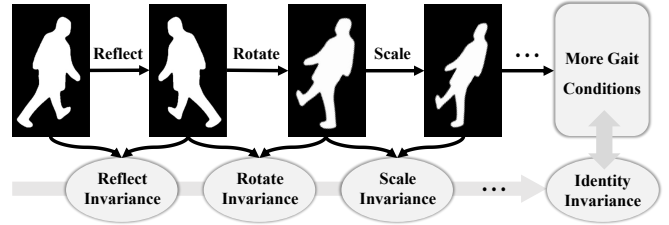}
        \caption{Different gait conditions are approximately associated with the combination of various geometric transformations, \eg, reflect, rotate, scale. Achieving invariance learning among them is an effective way to achieve identity invariance in gait recognition.}
        \label{fig:motivation}
    \end{figure}

    In ideal conditions, gait recognition strives to extract \textit{identity-invariant} features of an individual under various conditions.
    Over a long period, researchers have developed a series of gait datasets~\cite{zheng2022gait3d, zhu2021GREW,yu2006casiab,takemura2018oumvlp,li2023ccpg,shen2023lidargait} that encompass various gait conditions, including but not limited to normal walking, walking with different clothes, and multiple camera or walking views. 
    Subsequent gait models~\cite{chao2019gaitset,fan2020gaitpart,lin2021gaitgl,fan2023opengait} aim to extract \textit{common traits} of the same individual across various gait conditions in a data-driven manner for recognition.
    For example, during training, samples of the same individual across conditions are pulled closer, while those of different individuals are pushed apart.
    This paradigm has become mainstream and achieved remarkable performance~\cite{fan2023opengait,ma2024VPGait,wang2023dygait,dou2023gaitgci,wang2023HSTL,Huang2024GaitMoE,xiong2024CLTD}.

    However, current gait models largely rely on implicit training with existing \textit{limited} gait conditions. 
    Given the endless variety of conditions present in real-world scenarios, this reliance limits their ability to generalize to unseen conditions.
    To address this, an alternative strategy is to shift the focus to exploring the broader essence of understanding the explicit relations across different gait conditions.
    Once this essence is revealed, it becomes possible to establish connections among different gait conditions, thereby making generalization to broader conditions feasible.

    As an initial attempt, we propose a new perspective to approximately achieve gait recognition: variations in an individual’s gait under different conditions can be approximately viewed as a combination of geometric transformations. 
    For example, as in Fig.~\ref{fig:motivation},  
    (1) A 36-degree gait sequence can approximate a 144-degree sequence by horizontal reflection~\cite{yu2006casiab};
    (2) A lean-left gait sequence can be approximately generated by rotating an upright posture~\cite{zheng2022gait3d,li2023ccpg};
    (3) The thicker clothing effects can be approximately simulated by scaling~\cite{yu2006casiab,li2023ccpg}.
    In this way, more complex conditions can be viewed as a combination of geometric transformations. 
    Thus, \textbf{achieving identity invariance in gait recognition is approximately attainable through geometric invariance.}

    %
    Based on this new perspective, we focus on two primary questions: 
    (1) Which geometric transformations are most effective for modeling the variations in different gait conditions?
    (2) How can we achieve invariance across these geometric transformations in gait models?

    For the selection of geometric transformations, as an initial attempt, we choose three of the most common geometric transformations, \ie, \textbf{Reflect}, \textbf{Rotate}, \textbf{Scale}.
    Reflect approximates the changes in reflective walking directions. 
    Rotate simulates the variations in posture, like lean-left or lean-right. 
    Different clothing thicknesses can be considered as a type of Scale.
    %
    These transformations are prevalent in real-world gait scenarios, and their combination enables handling a wider range of walking conditions.
    Other geometric transformations are worth further exploration.

    To achieve geometric invariance, we propose a $\mathcal{R}$eflect-$\mathcal{R}$otate-$\mathcal{S}$cale invariance learning framework ($\mathcal{RRS}$\textbf{-Gait}). 
    The core idea is a two-stage process: it first learns approximately equivariant features, and then aggregates them into invariant features for recognition.
    For the first equivariance learning stage, it flexibly adjusts its convolution kernels based on the geometric transformation to achieve feature equivariance.
    For the second invariance learning stage, it introduces a pooling strategy to aggregate these equivariant features into a final invariant representation.
    Specifically, $\mathcal{RRS}$\textbf{-Gait} involves three equivariant modules and a invariant pooling layer:
    \textbf{(1) Reflect Equivariance Learning (ReEL)} module modifies the regular convolutional kernel to include half regular kernels and half \textit{reflected kernels} for feature extraction. The resulting two features are then aggregated along the corresponding channels into reflect-equivariant features. 
    \textbf{(2) Adaptive Rotate Equivariance Learning (RoEL)} module predicts rotation angle of each gait sequence and introduces \textit{rotated kernel} design to adapt to rotations for rotate equivariance.
    \textbf{(3) Multi-Scale Equivariance Learning (SEL)} module collects multi-scale features using \textit{various dilated kernels} and designs cross-channel and cross-scale feature interactions to enhance scale equivariance.
    \textbf{(4)} These three kinds of equivariant-aware features are respectively fed into each feature mapping stage with a \textit{pooling} operation to convert equivariant features into invariant ones.

    To summarize, the main contributions are as follows:
    \begin{itemize}
        \item We propose a new perspective to achieve gait recognition: the identity invariance under various gait conditions can be approximately achieved by geometric invariance. 
        \item As an initial attempt in this perspective, we explore three common geometric transformations (\ie, Reflect, Rotate, Scale) and their feasible equivariant kernel designs.   
        \item We propose $\mathcal{RRS}$-Gait, a $\mathcal{R}$eflect-$\mathcal{R}$otate-$\mathcal{S}$cale invariance learning framework, to achieve geometric equivariance and invariance for given geometric transformations.
        \item Extensive experiments on Gait3D, GREW, CCPG, and SUSTech1K show superior results across various conditions. For example, $\mathcal{RRS}$-Gait achieves 76.7\% and 81.0\% rank-1 in the challenging Gait3D and GREW.
    \end{itemize}


\section{Related Work} \label{sec:related_work}
    \subsection{Gait Recognition}
    Current gait methods can be primarily divided into two categories: model-based and appearance-based methods. 
    \textbf{Model-based methods}~\cite{fu2023gpgait,fan2024skeletongait,guo2023PAAICCV23,teepe2022gaitgraph2,zhu2023_3DMeshInference,shen2023lidargait,han2024freegait,huang2025watch,wang2023gait,li2022strong} utilize various estimation models to capture human structure (\eg, skeleton, mesh, and point cloud) as input, while \textbf{appearance-based methods}~\cite{chao2019gaitset,hou2020gaitGLN,lin2021gaitgl,wang2023gaitparsing,zheng2023parsinggait,liang2022gaitedge,ye2024biggait,wang2024qagait, wang2023hih, wang2023free,ye2025biggergait,qin2021rpnet,wang2025gaitc,xu2020cross,huang2022enhanced,yao2021collaborative,huang2024gaitdan,chen2025edinogait,zhang2025trackletgait} focus on human appearance (\eg, silhouette, parsing and RGB) for recognition.
    %
    Here we focus on silhouette-based methods, as they are most relevant to our work.

    Over a long period, silhouette-based methods often rely on temporal modeling, spatial modeling, or their combined modeling to capture discriminative features.
    For example, 
    \textbf{Spatial modeling}~\cite{chao2019gaitset,dou2023gaitgci,wang2023gaitparsing,zheng2023parsinggait,wang2023HSTL,hou2020gaitGLN,hou2021set,peng2024GLGait,qin2021rpnet,wang2025gait,li2019attentive} involves capturing global structural representations, local fine-grained details, and their interactions.
    \textbf{Temporal modeling}~\cite{fan2020gaitpart,lin2021gaitgl,huang2021CSTL,ma2023DANet,ma2024VPGait,wang2023dygait,fan2025opengait,huang2025learning,yang2025bridging,huang2022enhanced,deng2023human} focuses on inter-frame dynamics, including capturing gait cycles and mining various action sets (\eg, leg lifting, arm swinging), and multi-scale contextual temporal learning.
    These two strategies have achieved superior performance on in-the-lab gait datasets~\cite{yu2006casiab,takemura2018oumvlp}. 
    When extended to in-the-wild gait datasets~\cite{zheng2022gait3d,zhu2021GREW}, to enhance robustness against potential perturbations commonly present in real-world scenarios, recent works~\cite{fan2023opengait,ma2024VPGait,fan2025opengait} have widely adopted additional probabilistic data augmentation (\eg, $\sim$20\%, random flipping, random rotation, random erasing) and achieve further improvements.
    Despite their effectiveness, exploring more intrinsic solutions to establish connections across different gait conditions is of greater importance.
    
    %
    \subsection{Invariance Feature Learning}
    Invariance feature learning aims to extract features that remain stable under various transformations, making it widely used in computer vision, especially for classification with well-defined category labels~\cite{ye2021deep,shen2022comprehensiveSurvey,deng2009imagenet}.
    For example, the feature representations of a category under various conditions should be mapped to the same category label.
    The main strategies to achieve invariance learning include training with data augmentations~\cite{taylor2018improving, zoph2020learning}, pooling layers~\cite{laptev2016ti, zhang2019making}, or contrastive learning methods~\cite{chen2020simclr,he2020moco}.

    More recently, invariance feature learning can also be achieved by first implementing equivariance learning, where feature representations change predictably with input transformations, and then aggregating these equivariant features into invariant representations~\cite{weiler20183d,weiler2018learning,weiler2019general,weiler2023EquivariantAndCoordinateIndependentCNNs}.
    Equivariance is crucial for tasks like image segmentation~\cite{wang2022learning} and object detection~\cite{han2021redet,yu2022rotationally}.
    For example, the segmentation mask should change in response to transformations of objects in the image, whether they undergo translation, rotation, or other alterations.
    Equivariance learning often utilizes group theory~\cite{scott2012group} and has produced milestone works such as group convolutional networks~\cite{cohen2016group,romero2020attentive,weiler2019general} and rotationally equivariant CNNs~\cite{weiler20183d,weiler2018learning}.

\section{Preliminaries} \label{sec:method}
    \subsection{Why we need Equivariance and Invariance?}
        \begin{figure}[b]
            \centering
            \includegraphics[width=1.0\linewidth]{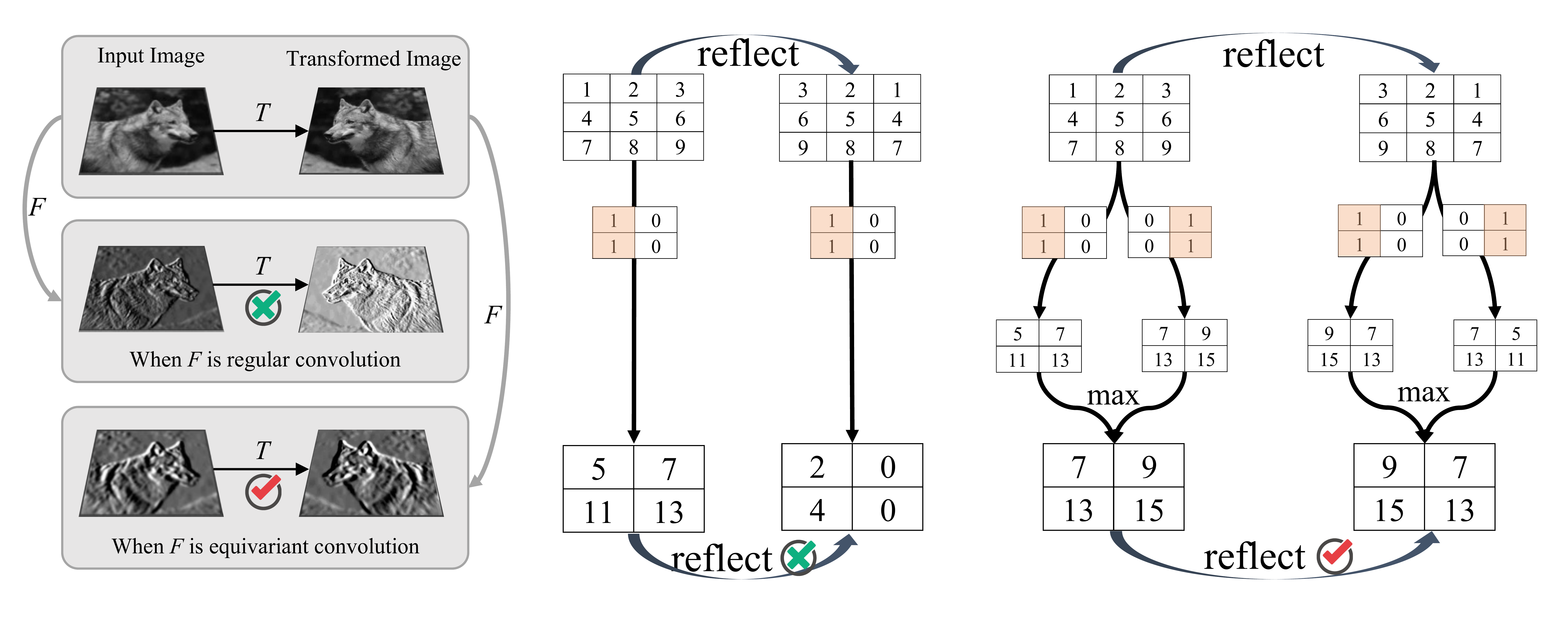}
            \caption{An example~\cite{weiler2023EquivariantAndCoordinateIndependentCNNs} illustrates equivariance in convolution. When applying a reflection transformation to the input image, the regular convolution kernel will produce non-equivariant feature maps, \ie, $\mathcal{F}(\mathcal{T}(\mathcal{X}))\neq\mathcal{T}(\mathcal{F}(\mathcal{X}))$. However, by introducing \textit{both regular kernel and its reflect kernel} for feature extraction, and applying max operation along their corresponding channels, we can obtain equivariant feature maps, \ie, $\mathcal{F}(\mathcal{T}(\mathcal{X}))=\mathcal{T}(\mathcal{F}(\mathcal{X}))$.}
            \label{fig:ills_equivariance}
        \end{figure}

        We first briefly review some preliminaries of transform-equivariance and transform-invariance~\cite{romero2020attentive,cesa2022ENequivariant}. 
        Given an image input $\mathcal{X}$ and a transformation $\mathcal{T}$, along with the network $\mathcal{F}$, we can say that: 

        %
        \begin{itemize}
            \item $\mathcal{F}$ is equivariant when $\mathcal{F}(\mathcal{T}(\mathcal{X}))=\mathcal{T}(\mathcal{F}(\mathcal{X}))$,
            \item $\mathcal{F}$ is invariant when $\mathcal{F}(\mathcal{T}(\mathcal{X}))=\mathcal{F}(\mathcal{X})$. 
        \end{itemize}

        The first property indicates the features extracted by the network $\mathcal{F}$ from input image $\mathcal{X}$ and transformed image $\mathcal{T}(\mathcal{X})$ also satisfy this transformation.
        As illustrated in Fig.~\ref{fig:ills_equivariance}, when taking the same image as input (differing only in reflect or not), an equivariant network can extract features consistent with this transformation. However, the regular convolution will produce distinct feature maps. This difference will be more pronounced as the convolution layers deepen, ultimately leading to distinct label predictions for the same image.

        The second property indicates the network $\mathcal{F}$ is unaffected by the transformation $\mathcal{T}$ applied to the input. When we obtain the above two equivariant feature maps, a simple spatial global pooling operation can ensure the pooled features are invariant.

        For gait recognition, \textit{assuming an individual under different gait conditions as a transformation, feature equivariance enables consistent responses in feature map~\cite{cesa2022ENequivariant}, thereby allowing extracting invariant features through pooling operation~\cite{weiler20183d,weiler2018learning,weiler2019general,weiler2023EquivariantAndCoordinateIndependentCNNs} to achieve identity invariance}.

    \subsection{How to achieve Equivariance and Invariance?}
    As illustrated in the right side of Fig.~\ref{fig:ills_equivariance}, recent studies in equivariant learning~\cite{weiler20183d,weiler2018learning,weiler2019general,weiler2023EquivariantAndCoordinateIndependentCNNs} have demonstrated that: 
    %
    \textit{Equivariant features can be achieved by adding extra convolution kernels (share parameters with regular kernel) based on the input transformation, and then aggregate their resulting feature maps along their corresponding channels into a compact representation.}
    Through explicitly integrating geometric transformation priors into kernel design, the model naturally ensures the produced features are in a predictable manner, \ie, equivariance.
    For example, Fig.~\ref{fig:ills_equivariance} shows that the reflection transformation equivariance can be achieved by introducing the reflected convolution kernel. 
    Additionally, some other works~\cite{rui2022ApproximatelyEquivariant,pu2023adaptiverotate} also explore \textit{flexibly adjusting the convolution kernel based on the input transformation can achieve approximate equivariance.}
    Finally, the equivariant features can be simply pooled into a vector to achieve invariance.
    
    \section{Proposed Method}
\begin{figure*}[t]
        \centering
        \includegraphics[width=1.0\linewidth]{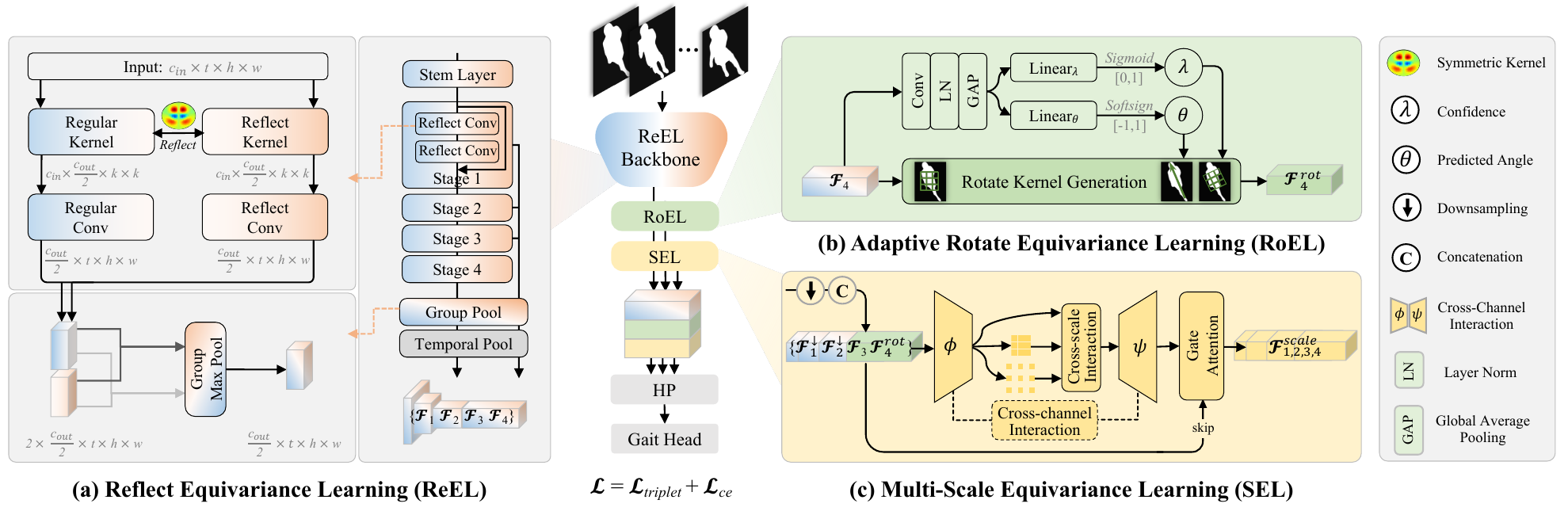}
        \caption{$\mathcal{R}$eflect-$\mathcal{R}$otate-$\mathcal{S}$cale invariance learning framework ($\mathcal{RRS}$-Gait) involves three equivariance learning modules and corresponding invariant pooling layers: (a) Reflect Equivariance Learning (ReEL) module, (b) Adaptive Rotate Equivariance Learning (RoEL) module, and (c) Multi-Scale Equivariance Learning (SEL) module respectively introduce \textit{reflect} kernels, \textit{rotate} kernels, and \textit{multiple dilated} kernels to achieve feature equivariance or approximate equivariance. These equivariant features are then fed into corresponding Horizontal Pooling (HP) layers with a global pooling operation to achieve final feature invariance or approximate invariance.}
        \label{fig:Pipeline_Overview}
    \end{figure*}

    In this paper, we select three geometric transformations (\ie, Reflect, Rotate, Scale) as an initial attempt to simulate different gait conditions.
    Subsequently, as in Fig.~\ref{fig:Pipeline_Overview}, \textbf{$\mathcal{RRS}$-Gait} proposes three equivariance learning modules and corresponding pooling layers for final invariance feature learning.

    \subsection{ReEL: Reflect Equivariance Learning} \label{sec:reflect}
        \noindent\textbf{$\divideontimes$ Necessity.} 
        Various walking directions of an individual make gait sequences involving multiple walking views. 
        To improve feature consistency issues among them, researchers often construct rich cross-view conditions within in-the-lab settings~\cite{yu2006casiab,takemura2018oumvlp}. 
        When shifting to in-the-wild settings, the uncontrolled and limited camera views will bring another sparse-view issue~\cite{zheng2022gait3d,zhu2021GREW,wang2023sparseview}.
        Based on this, Reflect Equivariance Learning (ReEL) is beneficial and necessary to address the above issues from two aspects:
        \begin{itemize}
            \item For in-the-lab case, ReEL helps bring closer the gait sequences from reflectively related views. 
            \item For in-the-wild case, ReEL helps the gait model adapt to the absent reflective view.
        \end{itemize}

        \vspace{0.1cm}
        \noindent\textbf{$\divideontimes$ Solution.} 
        %
        As shown in the left part of Fig.~\ref{fig:Pipeline_Overview}, we modify the regular convolution kernel to include \textit{half regular kernels} and \textit{half reflected kernels} for feature extraction. 
        Note that kernel reflection is not equal to feature map reflection. Like the Sobel kernel design in edge detection~\cite{farid1997optimally,farid2004differentiation}, the horizontal kernel and vertical kernel capture different features, \ie, horizontal edges and vertical edges.
        Then the extracted features with regular and reflected kernels will be aggregated along the corresponding channels (named Group Pool) to obtain a reflect-equivariant feature.
        In this case, as shown in Fig.~\ref{fig:ills_equivariance}, whether the input is the original image or the reflected image, they will ultimately be aggregated into this unified feature to achieve reflect equivariance.

    \subsection{RoEL: Adaptive Rotate Equivariance Learning}
        \noindent\textbf{$\divideontimes$ Necessity.} 
        Considering both indoor and outdoor gait data are usually collected from monitoring or other uncontrolled scenarios, different camera angles will cause various left or right rotation in different gait sequences~\cite{zheng2022gait3d,zhu2021GREW}.
        However, 
        
        \begin{itemize}
            %
            \item Regular CNNs primarily focus on translation invariance but struggle with rotation variances~\cite{han2021redet,chidester2018rotation,lee2024fred}.
            \item Researchers~\cite{fan2023opengait} often use probabilistic ($\sim$20\%) rotate augmentation with a small range of angle (\eg, $\pm10^{\circ}$) to adapt to rotated samples, but this \emph{pre-defined} strategy relies on the empirical setting. A large probability or angle may degrade the performance ~\cite{li2023ccpg,fan2023opengait}.
        \end{itemize}

        Thus, it is necessary to \textit{adaptively mitigate rotation variances}, while \textit{covering a broader range of samples}.

        \vspace{0.2cm}
        \noindent\textbf{$\divideontimes$ Difficulty.}
        If following reflect kernel design in Sec.~\ref{sec:reflect}, adding \emph{rotate-equivariant} kernels seems straightforward. However, this is not efficient from two aspects:
        
        \begin{itemize}
            \item \emph{About equivariance.} Unlike reflection that has fixed and limited \{2\} transformations (\ie, reflect or not), rotation is continuous and infinite \{0$^{\circ}$-360$^{\circ}$\}. Only by involving all angles to conduct rotated kernels can we achieve the strict equivariance learning. However, in practice, we can only choose discrete sampling angles, which will lead to approximate equivariance.
            \item \emph{About efficiency.} Although approximate equivariance may also be acceptable, each rotated angle requires doubling the number of kernels, exponentially increasing computational cost. However, infinitely halving kernels like Sec.~\ref{sec:reflect} to maintain efficiency is not feasible.
        \end{itemize}

        \noindent\textbf{$\divideontimes$ Solution.} 
        As shown in Fig.~\ref{fig:Pipeline_Overview}(b), to search for a trade-off between effectiveness and efficiency, motivated by recent advancements in convolutional research~\cite{chen2024frequencydilated,pu2023adaptiverotate}, we introduce \textbf{Adaptive Rotate Equivariance Learning (RoEL)} to adaptively predict the rotated angle of each gait sequence in a data-driven manner, and then adaptively perform rotate convolution for approximate rotate equivariance. All involve the following five steps:

        \vspace{0.1cm}
        \noindent\textbf{{Step 1: Feature Selection.}} 
        Accurate prediction of gait rotated angle requires reliable features. We empirically select last-stage features ($\mathcal{F}_{4}^{'}\in \mathbb{R}^{C_{4}\times T\times \frac{H}{4}\times \frac{W}{4}}$) for this purpose.

        \vspace{0.1cm}
        \noindent\textbf{{Step 2: Temporal Aggregation.}} 
        Since gait sequence is often captured by the same camera in a short time, they largely share a similar rotated angle. 
        Thus, we introduce \emph{Temporal Max Pooling} to aggregate the maximum response of temporal features, which is also suitable for mitigating the impact of abnormal or redundant frames and reducing calculations.
        %
        \begin{equation}\label{eq:TA}
            \mathcal{F}_{4} = \texttt{TemporalAgg}(\mathcal{F}_{4}^{'}), \hspace{0.1cm} \mathcal{F}_{4}\in \mathbb{R}^{C_{4}\times \frac{H}{4}\times \frac{W}{4}}.
        \end{equation}
        
        \noindent\textbf{{Step 3: Rotated Angle Prediction ($\theta$) and Confidence ($\lambda$).}} 
        As in Fig.~\ref{fig:Pipeline_Overview}, the output feature ($\mathcal{F}_{4}\in \mathbb{R}^{C_{4}\times \frac{H}{4}\times \frac{W}{4}}$) is first processed by a convolution, layer normalization, and ReLU activation to effectively extract spatial features and ensure feature stability. 
        Then, this feature undergoes Global Average Pooling (${\rm GAP}$) to eliminate spatial dimensions ($\mathbb{R}^{C_{4}\times 1}$).
        Finally, it will be fed into two \emph{independent} branches to simultaneously learn the rotated angle ($\theta$) and confidence ($\lambda$).
        Each branch includes a linear layer ($C_{4}$$\rightarrow$$1$) and an activation function. 
        The difference between them is the selection of activation function: the rotated angle can be positive or negative, so it uses Softsign~\cite{glorot2010softsign} to output in the range $[-1, 1]$, while the confidence branch uses Sigmoid~\cite{sharma2017activation} to output in the range $[0, 1]$. 
        Additionally, the angle will not be multiplied by 360$^{\circ}$ but by a limited angle (\eg, $\theta_{limit}=20^{\circ}/30^{\circ}/40^{\circ}/50^{\circ}$) since human body usually leans within a limited range.
        The overall are as follows:
        %
        \begin{equation}
            \fontsize{9.5pt}{10pt}\selectfont
            \mathcal{F}^{rot}_{init} = \texttt{GAP}(\texttt{ReLU}(\texttt{LayerNorm}(\texttt{Conv2d}(\mathcal{F}_{4})))),
        \end{equation}
        \begin{equation}\label{eq:theta_limit}
            \theta = \texttt{Softsign}(\texttt{Linear}_{\theta}(\mathcal{F}^{rot}_{init})) \cdot \theta_{limit}, \hspace{0.1cm} \theta \in \mathbb{R}^{1},
        \end{equation}
        \begin{equation}
            \lambda = \texttt{Sigmoid}(\texttt{Linear}_{\lambda}(\mathcal{F}^{rot}_{init})), \hspace{0.1cm} \lambda \in \mathbb{R}^{1}.
        \end{equation}

        \noindent\textbf{{Step 4: Adaptive Rotated Kernel Generation ($\mathcal{K}$$\rightarrow$$\mathcal{K}_{\theta}$).}}
        Based on the regular kernel ($\mathcal{K}$) and the predicted angle ($\theta$), we construct the adaptive rotation matrix and utilize bilinear interpolation to generate the final rotated kernel ($\mathcal{K}_{\theta}$).
        \begin{equation}
            \mathcal{K}_{\theta} = \texttt{Rotate}(\mathcal{K}, \theta), \hspace{0.1cm} \{\mathcal{K}, \mathcal{K}_{\theta}\}\in \mathbb{R}^{C_{4}\times C_{4}\times k\times k},
        \end{equation}
        
        \noindent where $C_{4}$ and $k$ are input/output channels and kernel size.
        
        \noindent\textbf{{Step 5: Adaptive Rotate Convolution.}}
        The generated rotated kernel ($\mathcal{K}_{\theta}$) replaces regular kernel ($\mathcal{K}$) for adaptive rotation convolution with confidence ($\lambda$) as follows:
        \begin{equation}
            \mathcal{F}_{4}^{rot} = \lambda \cdot \mathcal{K}_{\theta}(\mathcal{F}_{4}), \hspace{0.2cm} \{\mathcal{F}_{4}^{rot},\mathcal{F}_{4}\}\in \mathbb{R}^{C_{4}\times \frac{H}{4}\times \frac{W}{4}}.
        \end{equation}


    \subsection{SEL: Multi-Scale Equivariance Learning}
        \noindent\textbf{$\divideontimes$ Necessity.}
        Similar to Reflect and Rotate, the Scale variance is also common in gait field. 
        Wearing a coat or skirt will cause scale variations in the shape of human body.
        %
        %
        This inspires researchers to construct corresponding gait conditions (\eg, cloth-changing~\cite{li2023ccpg}) to simulate scale variance, but it remains highly challenging and has not achieved satisfactory results.
        
        %
        \noindent\textbf{$\divideontimes$ Difficulty.}
        Similar to Rotate, Scale transformations are also continuous and infinite \{$\infty$\}.
        How to adaptively adapt to various scale variations while maintaining efficiency is just as urgent as addressing Rotate variances.
        
        %
        \noindent\textbf{$\divideontimes$ Solution.}
        To simulate scale transformations and mitigate scale variances, we introduce \textbf{Multi-Scale Equivariance Learning (SEL)}. It collects multi-scale features in each layer as the initial input and designs \textit{multiple dilated kernels} to further simulate scale variances, facilitating the following multi-scale interactions to mitigate scale variance and achieve approximate scale equivariance, involving:

        %
        \noindent\textbf{{Step 1: Multi-scale Feature Selection.}}
        To obtain multi-scale gait representations, we select the output of each layer in backbone ($\mathcal{F}_{1}, \mathcal{F}_{2}, \mathcal{F}_{3}, \mathcal{F}_{4}^{rot}$), as they involve varying spatial resolutions and receptive fields due to subsampling layers and convolutions of different depths in the backbone.

        \noindent\textbf{{Step 2: Feature Initialization.}}
        To ensure efficiency, we assume scale variances of frames within a gait sequence are similar since they are always captured by the same camera in a short time. We adopt \textit{Temporal Max Pooling} as Eq.~(\ref{eq:TA}) to aggregate the maximum response of temporal features.
        Then, all of them are downsampled to the same resolution ($\mathbb{R}^{C_{i}\times \frac{H}{4}\times \frac{W}{4}}$) based on bilinear interplation and concatenated along channel dimension to obtain initial scale-orient feature ($\mathcal{F}_{init}^{scale}\in \mathbb{R}^{(C_1+C_2+C_3+C_4)\times \frac{H}{4}\times \frac{W}{4}}$) as follows. 
        \begin{equation}
            \begin{aligned}
                \mathcal{F}_{init}^{scale} &= \underset{C}{\texttt{Concat}}(\mathcal{F}^{\downarrow}_{1},\mathcal{F}^{\downarrow}_{2},\mathcal{F}_{3},\mathcal{F}^{rot}_{4}).
            \end{aligned}
        \end{equation}
        
        %
        \noindent\textbf{Step 3: Cross-Channel Interaction Module.}
        Based on the initial scale-orient features ($\mathcal{F}_{init}^{scale}$), we simply introduce a convolution layer with batch normalization and a nonlinear layer to implement channel reduction, facilitating cross-channel interaction to mix multi-scale features.
        %
        \begin{equation}\label{eq:CCIM_1}
            \phi: \ \ \mathcal{F}_{c}^{scale} = \texttt{ReLU}(\texttt{BN}(\underset{C\rightarrow C/r}{\texttt{Conv2d}}(\mathcal{F}_{init}^{scale}))).
        \end{equation}
        \normalsize

        %
        \noindent\textbf{Step 4: Cross-Scale Interaction Module.}
        To better capture multiple scales for spatial interaction, we further introduce multi-scale convolutions based on: {\footnotesize \circled{s1}} no operation as the original feature, {\footnotesize \circled{s2}} kernel=3, and {\footnotesize \circled{s3}} kernel=5, to obtain three spatial scales, \ie, $\mathcal{F}_{s_1}^{scale}$, $\mathcal{F}_{s_2}^{scale}$, and $\mathcal{F}_{s_3}^{scale}$.

        Subsequently, we flatten the spatial dimension of these three scales and apply vanilla self-attention~\cite{vaswani2017transformer} for multi-scale interaction.
        Specifically, the original feature ({\footnotesize \circled{s1}}) serve as the query and the other two ({\footnotesize \circled{s2}}{\footnotesize \circled{s3}}) are regarded as key and value to achieve cross-scale attention, followed by the vanilla Feed Forward (${\rm FFN}$) layer~\cite{vaswani2017transformer}.
        \begin{equation}
            \begin{aligned}
                q=W_{q}\hspace{-0.05cm}\cdot\hspace{-0.05cm}\mathcal{F}_{s_1}^{scale}, k=W_{k}\hspace{-0.05cm}\cdot\hspace{-0.05cm}\mathcal{F}_{s_2}^{scale}, v=W_{v}\hspace{-0.05cm}\cdot\hspace{-0.05cm}\mathcal{F}_{s_3}^{scale}, \\
                \mathcal{F}^{scale}_{s} = \texttt{FFN}(\texttt{Softmax}(\frac{q\cdot k^{T}}{\sqrt{C/r}})\cdot v),
            \end{aligned}
        \end{equation}
        where $W_{q}$, $W_{k}$, and $W_{v}$ are weights of linear layer as \cite{vaswani2017transformer}.
        
        %
        
        \noindent\textbf{Step 5: Gate Attention.}
        Finally, we introduce gate attention with residual connection to mitigate early training instability and improve optimization. 
        Specifically, we first expand the channels of $\mathcal{F}^{scale}_{s}$ to the original size ($C/r\rightarrow C$), which can be viewed as a second cross-channel interaction. 
        \begin{equation}\label{eq:CCIM_2}
            \psi: \ \ \mathcal{F}_{s}^{scale} \leftarrow \texttt{ReLU}(\texttt{BN}(\underset{C/r\rightarrow C}{\texttt{Conv2d}}(\mathcal{F}_{s}^{scale}))),
        \end{equation}
        
        
        %
        Then, it will be split into four outputs with the same size of $\mathcal{F}_{i} \in \{\mathcal{F}_{1}^{\downarrow}$,$\mathcal{F}_{2}^{\downarrow}$,$\mathcal{F}_{3}^{}$,$\mathcal{F}_{4}^{rot}\}$ along the channel dimension.
        The overall gate attention is as follows:
        \begin{equation}
            \mathcal{F}^{scale}_{i} = \texttt{Sigmoid}(\texttt{BN}(\underset{\texttt{k=1}}{\texttt{Conv2d}_{i}}(\mathcal{F}_{s,i}^{scale})))\cdot \mathcal{F}_{s,i}^{scale} + \mathcal{F}_{i},
        \end{equation}
        
        \noindent where $\mathcal{F}^{scale}_{i}$ indicates the scaled features, $i=1,2,3,4$.

    \begin{table*}
        \centering
        \setlength{\tabcolsep}{2.5mm}
        \caption{Implementation details on both indoor and outdoor datasets. \#IDs and \#Seqs indicate the total subjects and sequences. \#lr and \#wd are the initial learning rate and weight decay for the optimizer.}
        \label{tab:ImplementationDetails}
        \resizebox{1.0\linewidth}{!}{
            \begin{tabular}{c|c|cc|cc|c|c|c|c}
            \toprule
            \multirow{2}{*}{Environments} & \multirow{2}{*}{Datasets} & \multicolumn{2}{c|}{Train Phase} & \multicolumn{2}{c|}{Test Phase} & \multirow{2}{*}{\begin{tabular}[c]{@{}c@{}}Batch\\      (P, K)\end{tabular}} & \multirow{2}{*}{Optimizer} & \multirow{2}{*}{Milestones} & \multirow{2}{*}{\begin{tabular}[c]{@{}c@{}}Total \\      Iters\end{tabular}} \\ \cline{3-6}
            &  & \#IDs & \#Seqs & \#IDs & \#Seqs &  &  &  &  \\
            \midrule
            \multirow{3}{*}{Outdoor} & Gait3D~\cite{zheng2022gait3d} & 3,000 & 18,940 & 1,000 & 6,369 & (32, 4) & \multirow{4}{*}{\begin{tabular}[c]{@{}c@{}}SGD\\      \#lr=0.1\\      \#wd=0.0005\\      momentum=0.9\end{tabular}} & (20K,40K,50K) & 60K \\
            & GREW~\cite{zhu2021GREW} & 20,000 & 102,887 & 6,000 & 24,000 & (32, 4) &  & (80K,120K,150K) & 180K \\
            & SUSTech1K~\cite{shen2023lidargait} & 250 & 6,011 & 800 & 19,228 & (8, 8) &  & (20K,30K) & 40K \\
            Indoor \& Outdoor & CCPG~\cite{li2023ccpg} & 100 & 8,187 & 100 & 8,095 & (8, 16) &  & (20K,40K,50K) & 60K \\
            \bottomrule
            \end{tabular}
        }
    \end{table*}
    
    \begin{table}[t]
        \centering
        \caption{Architecture of $\mathcal{RRS}$-Gait. $\mathrm{ReflectConv}$ indicates Reflect Convolution to replace regular convolution. ``T'', ``C'' denote sequence length and output channels. Stride and residual are omitted for simplicity. C=32$_{\times 2}$ means 32 channels are from regular kernels and another 32 from reflected kernels. ``$l_{i}$'' is the number of layers in each stage. The \colorbox[HTML]{ECF4FF}{color} highlights our modifications.}
        \label{tab:model_main_architecture}
        \setlength{\tabcolsep}{1.5mm}
        \renewcommand\arraystretch{0.95}
        \resizebox{1.0\linewidth}{!}{
            \begin{tabular}{c|c|c}
                \toprule
                \textbf{Layer} & \textbf{Output} & \textbf{$\mathcal{RRS}$-Gait} \\
                \midrule
                Stem & $T\times H\times W$ & \colorbox[HTML]{ECF4FF}{$\begin{array}{c} {\rm ReflectConv2D, Kernel=3, C=32_{\times 2}} \end{array}$} \\
                \midrule
                \multirow{2}{*}{\vspace*{0.5cm}Stage1} & \multirow{2}*{\vspace*{0.5cm}$T\times H\times W$} & \colorbox[HTML]{ECF4FF}{$\left[\!\!\! \begin{array}{c} {\rm ReflectConv2D, Kernel=3, C=32_{\times 2}} \\ {\rm ReflectConv2D, Kernel=3, C=32_{\times 2}} \end{array} \!\!\! \right ] \!\!\times\!  l_1$} \\
                \midrule
                \multirow{2}*{\vspace*{0.5cm}Stage 2} & \multirow{2}*{\vspace*{0.5cm}$T\times \frac{H}{2} \times \frac{W}{2}$} & \colorbox[HTML]{ECF4FF}{$\left[\!\!\! \begin{array}{c} {\rm ReflectConv, Kernel=3, C=64_{\times 2}} \\ {\rm ReflectConv, Kernel=3, C=64_{\times 2}} \end{array} \!\!\! \right ] \!\!\times\!  l_2$} \\
                \midrule
                \multirow{2}*{\vspace*{0.5cm}Stage 3} & \multirow{2}*{\vspace*{0.5cm}$T\times \frac{H}{4}\times \frac{W}{4}$} & \colorbox[HTML]{ECF4FF}{$\left[\!\!\! \begin{array}{c} {\rm ReflectConv, Kernel=3, C=128_{\times 2}} \\ {\rm ReflectConv, Kernel=3, C=128_{\times 2}} \end{array} \!\!\! \right ] \!\!\times\!  l_3$} \\
                \midrule
                \multirow{2}*{\vspace*{0.5cm}Stage 4} & \multirow{2}*{\vspace*{0.5cm}$T\times \frac{H}{4}\times \frac{W}{4}$} & \colorbox[HTML]{ECF4FF}{$\left[\!\!\! \begin{array}{c} {\rm ReflectConv, Kernel=3, C=256_{\times 2}} \\ {\rm ReflectConv, Kernel=3, C=256_{\times 2}} \end{array} \!\!\! \right ] \!\!\times\! l_4$} \\
                \midrule
                \rowcolor[HTML]{ECF4FF} 
                GPool & $T\times \frac{H}{4}\times \frac{W}{4}$ & Group Pool (C $=256_{\times 2}\rightarrow 256$)\\
                \midrule
                TP & $1\times \frac{H}{4}\times \frac{W}{4}$ & \multicolumn{1}{c}{Temporal Max Pooling ($T \rightarrow 1$)} \\
                \midrule
                \rowcolor[HTML]{ECF4FF} 
                RoEL & $1\times \frac{H}{4}\times \frac{W}{4}$ & Adaptive Rotate Equivariance Learning \\
                \midrule
                \rowcolor[HTML]{ECF4FF} 
                SEL & $1\times \frac{H}{4}\times \frac{W}{4}$ & \multicolumn{1}{c}{Multi-Scale Equivariance Learning} \\
                \midrule
                HP & $1\times 3P$ & \multicolumn{1}{c}{Horizontal Pooling~\cite{chao2019gaitset}} \\
                \midrule
                \multicolumn{1}{c|}{Head} & $1\times 3P$ & \multicolumn{1}{c}{Seprate Fully Connected Layers \& BNNeck~\cite{chao2019gaitset}} \\
                \bottomrule
            \end{tabular}
            }
    \end{table}

    \begin{table}[t]
        \centering
        \setlength{\tabcolsep}{1.8mm}
        \caption{Single-domain evaluation results on Gait3D~\cite{zheng2022gait3d} and GREW~\cite{zhu2021GREW} datasets with Rank, mAP, and mINP accuracy (\%). The \textbf{Bold} indicates best result.} 
        \label{tab:Gait3D_GREW_Results}
        \resizebox{1.0\linewidth}{!}{
            \begin{tabular}{cc|ccc|cccc}
                \toprule
                \multicolumn{1}{c|}{\multirow{2}{*}{Method}} & \multirow{2}{*}{Venue} & \multicolumn{3}{c|}{Gait3D~\cite{zheng2022gait3d}} & \multicolumn{2}{c}{GREW~\cite{zhu2021GREW}} \\ 
                \cmidrule{3-7} 
                \multicolumn{1}{c|}{} &  & R-1 & R-5 & mAP & R-1 & R-5 \\ 
                \midrule
                \multicolumn{1}{c|}{GaitGraph2~\cite{teepe2022gaitgraph2}} & CVPRW'22 & 11.2 & - & - & 34.8 & - \\
                \multicolumn{1}{c|}{GaitTR~\cite{zhang2023GaitTR}} & ES'23 & 7.2 & - & - & 48.6 & - \\
                \multicolumn{1}{c|}{PAA~\cite{guo2023PAAICCV23}} & ICCV'23 & 38.9 & - & - & 38.7 & 62.1 \\
                \multicolumn{1}{c|}{SkeletonGait~\cite{fan2024skeletongait}} & AAAI'24 & 38.1 & 56.7 & 28.9 & 77.4 & 87.9 \\
                \midrule
                \multicolumn{1}{c|}{GaitSet~\cite{chao2019gaitset}} & AAAI'19 & 36.7 & 58.3 & 30.0 & 46.3 & 63.6 \\
                \multicolumn{1}{c|}{GaitPart~\cite{fan2020gaitpart}} & CVPR'20 & 28.2 & 47.6 & 21.6 & 44.0 & 60.7 \\
                \multicolumn{1}{c|}{GaitGL~\cite{lin2021gaitgl}} & ICCV'21 & 29.7 & 48.5 & 22.3 & 47.3 & 63.6 \\
                \multicolumn{1}{c|}{CSTL~\cite{huang2021CSTL}} & ICCV'21 & 11.7 & 19.2 & 5.6 & 50.6 & 65.9 \\
                \multicolumn{1}{c|}{DANet~\cite{ma2023DANet}} & CVPR'23 & 48.0 & 69.7 & - & - & - \\
                \multicolumn{1}{c|}{GaitGCI~\cite{dou2023gaitgci}} & CVPR'23 & 50.3 & 68.5 & 39.5 & 68.5 & 80.8 \\
                \multicolumn{1}{c|}{GaitBase~\cite{fan2023opengait}} & CVPR'23 & 64.6 & - & 55.2 & 60.1 & - \\
                \multicolumn{1}{c|}{HSTL~\cite{wang2023HSTL}} & ICCV'23 & 61.3 & 76.3 & 55.5 & 62.7 & 76.6 \\
                \multicolumn{1}{c|}{DyGait~\cite{wang2023dygait}} & ICCV'23 & 66.3 & 80.8 & 56.4 & 71.4 & 83.2 \\
                \multicolumn{1}{c|}{GaitSSB~\cite{fan2023gaitssb}} & TPAMI'23 & 63.6 & - & - & 61.7 & - \\
                \multicolumn{1}{c|}{CLASH~\cite{dou2024clash}} & TIP'24 & 52.4 & 69.2 & 40.2 & 67.0 & 78.9 \\
                \multicolumn{1}{c|}{HiH-S~\cite{wang2023hih}} & ArXiv'23 & 72.4 & 86.9 & 64.4 & 72.5 & 83.6 \\
                \multicolumn{1}{c|}{QAGait~\cite{wang2024qagait}} & AAAI'24 & 67.0 & 81.5 & 56.5 & 59.1 & 74.0 \\
                \multicolumn{1}{c|}{DeepGaitV2~\cite{fan2025opengait}} & ArXiv'23 & 74.4 & 88.0 & 65.8 & 77.7 & 88.9 \\
                \multicolumn{1}{c|}{VPNet-L~\cite{ma2024VPGait}} & CVPR'24 & 75.4 & 87.1 & - & 80.0 & 89.4 \\
                \multicolumn{1}{c|}{CLTD~\cite{xiong2024CLTD}} & ECCV'24 & 69.7 & 85.2 & - & 78.0 & 87.8 \\
                \multicolumn{1}{c|}{FreeLunch~\cite{wang2023free}} & ECCV'24 & 70.1 & - & 61.9 & 65.5 & 78.7 \\
                \multicolumn{1}{c|}{GaitMoE~\cite{Huang2024GaitMoE}} & ECCV'24 & 73.7 & - & 66.2 & 79.6 & 89.1 \\ 
                
                \midrule
                \rowcolor[HTML]{ECF4FF} 
                \multicolumn{2}{c|}{{Baseline-L (Ours)}} & 74.4 & 88.0 & 65.8 & 79.2 & 89.2 \\ 
                \rowcolor[HTML]{ECF4FF} 
                \multicolumn{2}{c|}{{\textbf{$\mathcal{RRS}$-Gait-L (Ours)}}} & \textbf{76.7} & \textbf{89.9} & \textbf{69.6} & \textbf{81.0} & \textbf{89.9} \\ 
                
                \bottomrule
            \end{tabular}
        }
    \end{table}

    \begin{table*}[t]
        \centering
        \setlength{\tabcolsep}{4.0mm}
        \caption{Comparison results on CCPG~\cite{li2023ccpg} dataset based on Gait and ReID evaluation protocols with Rank-1 accuracy (\%). The \textbf{Bold} indicates best result.}
        \label{tab:CCPG_Results}
        \resizebox{0.95\linewidth}{!}{
            \begin{tabular}{cc|cccc|c|cccc|c}
            \toprule
            \multicolumn{1}{c|}{} &  & \multicolumn{5}{c|}{Gait Evaluation Protocol (R-1)} & \multicolumn{5}{c}{ReID Evaluation Protocol (R-1)} \\ 
            \cmidrule{3-12} 
            \multicolumn{1}{c|}{\multirow{-2}{*}{Method}} & \multirow{-2}{*}{Venue} & CL & UP & DN & BG & Mean & CL & UP & DN & BG & Mean \\ 
            \midrule
            \multicolumn{1}{c|}{GaitGraph2~\cite{teepe2022gaitgraph2}} & CVPRW’22 & 5.0 & 5.3 & 5.8 & 6.2 & 5.6 & 5.0 & 5.7 & 7.3 & 8.8 & 6.7 \\
            \multicolumn{1}{c|}{GaitTR~\cite{zhang2023GaitTR}} & ES’23 & 15.7 & 18.3 & 18.5 & 17.5 & 17.5 & 24.3 & 28.7 & 31.1 & 28.1 & 28.1 \\
            \multicolumn{1}{c|}{SkeletonGait~\cite{fan2024skeletongait}} & AAAI'24 & 40.4 & 48.5 & 53.0 & 61.7 & 50.9 & 52.4 & 65.4 & 72.8 & 80.9 & 67.9 \\
            \midrule
            \multicolumn{1}{c|}{GaitSet~\cite{chao2019gaitset}} & AAAI'19 & 60.2 & 65.2 & 65.1 & 68.5 & 64.8 & 77.5 & 85.0 & 82.9 & 87.5 & 83.2 \\
            \multicolumn{1}{c|}{GaitPart~\cite{fan2020gaitpart}} & CVPR'20 & 64.3 & 67.8 & 68.6 & 71.7 & 68.1 & 79.2 & 85.3 & 86.5 & 88.0 & 84.8 \\
            \multicolumn{1}{c|}{AUG-OGBase~\cite{li2023ccpg}} & CVPR'23 & 52.1 & 57.3 & 60.1 & 63.3 & 58.2 & 70.2 & 76.9 & 80.4 & 83.4 & 77.7 \\
            \multicolumn{1}{c|}{GaitBase~\cite{fan2023opengait}} & CVPR'23 & 71.6 & 75.0 & 76.8 & 78.6 & 75.5 & 88.5 & 92.7 & 93.4 & 93.2 & 92.0 \\
            \midrule
            \rowcolor[HTML]{ECF4FF} 
            \multicolumn{2}{c|}{\cellcolor[HTML]{ECF4FF}{Baseline-S (Ours)}} & 71.8 & 75.8 & 77.1 & 79.0 & 75.9 & 88.5 & 92.9 & 93.0 & 93.2 & 91.9 \\
            \rowcolor[HTML]{ECF4FF} 
            \multicolumn{2}{c|}{\cellcolor[HTML]{ECF4FF}{\textbf{$\mathcal{RRS}$-Gait-S (Ours)}}} & \textbf{78.4} & \textbf{81.8} & \textbf{81.9} & \textbf{85.7} & \textbf{82.0} & \textbf{91.8} & \textbf{96.5} & \textbf{94.4} & \textbf{96.2} & \textbf{94.7} \\
            \bottomrule
            \end{tabular}
        }
    \end{table*}

    \begin{table*}[t]
        \centering
        \setlength{\tabcolsep}{2.5mm}
        \caption{Comparison results on SUSTech1K~\cite{shen2023lidargait} dataset with various conditions under Rank-1 and Rank-5 accuracy (\%). The \textbf{Bold} indicates best result.}
        \label{tab:SUSTech1K_results}
        \resizebox{0.95\linewidth}{!}{
            \begin{tabular}{c|c|cccccccc|cc}
            \toprule
             &  & \multicolumn{8}{c|}{Probe Sequence (R-1)} & \multicolumn{2}{c}{Overall} \\ 
             \cmidrule{3-12} 
            \multirow{-2}{*}{Method} & \multirow{-2}{*}{Venue} & Normal & Bag & Clothing & Carrying & Umberalla & Uniform & Occlusion & Night & R-1 & R-5 \\ 
            \midrule
            GaitGraph2~\cite{teepe2022gaitgraph2} & CVPRW'22 & 22.2 & 18.2 & 6.8 & 18.6 & 13.4 & 19.2 & 27.3 & 16.4 & 18.6 & 40.5 \\
            GaitTR~\cite{zhang2023GaitTR} & ES'23 & 33.3 & 31.5 & 21.0 & 30.4 & 22.7 & 34.6 & 44.9 & 23.5 & 30.8 & 56.0 \\
            SkeletonGait~\cite{fan2024skeletongait} & AAAI'24 & 67.9 & 63.5 & 36.5 & 61.6 & 58.1 & 67.2 & 79.1 & 50.1 & 63.0 & 83.5 \\
            \midrule
            GaitSet~\cite{chao2019gaitset} & AAAI'19 & 69.1 & 68.2 & 37.4 & 65.0 & 63.1 & 61.0 & 67.2 & 23.0 & 65.0 & 84.8 \\
            GaitPart~\cite{fan2020gaitpart} & CVPR'20 & 62.2 & 62.8 & 33.1 & 59.5 & 57.2 & 54.8 & 57.2 & 21.7 & 59.2 & 80.8 \\
            GiatGL~\cite{lin2021gaitgl} & ICCV'21 & 67.1 & 66.2 & 35.9 & 63.3 & 61.6 & 58.1 & 66.6 & 17.9 & 63.1 & 82.8 \\ 
            GaitBase~\cite{fan2023opengait} & CVPR'23 & 81.5 & 77.5 & 49.6 & 75.8 & 75.5 & 76.7 & 81.4 & 25.9 & 76.1 & 89.4 \\
            \midrule
            \rowcolor[HTML]{ECF4FF} 
            \multicolumn{2}{c|}{\cellcolor[HTML]{ECF4FF}{Baseline-S (Ours)}} & 81.5 & 77.8 & 49.7 & 75.6 & 75.3 & 75.9 & 81.8 & 25.9 & 76.0 & 89.5 \\
            \rowcolor[HTML]{ECF4FF} 
            \multicolumn{2}{c|}{\cellcolor[HTML]{ECF4FF}{\textbf{$\mathcal{RRS}$-Gait-S (Ours)}}} & \textbf{85.7} & \textbf{82.2} & \textbf{52.3} & \textbf{79.7} & \textbf{81.4} & \textbf{81.0} & \textbf{87.3} & \textbf{29.0} & \textbf{80.3} & \textbf{92.0} \\
             \bottomrule
            \end{tabular}
        }
    \end{table*}

    \subsection{Invariant Representation and Loss Function}
        \noindent\textbf{Invariant Representation.}
        To this end, we obtain three equivariant or approximately equivariant features: Reflect ($\mathcal{F}_{4}$), Rotate ($\mathcal{F}_{4}^{rot}$), and Scale ($\mathcal{F}_{1,2,3,4}^{scale}$). 
        Following recent works~\cite{wang2023gaitparsing,zheng2023parsinggait,wang2023HSTL}, as in Table~\ref{tab:model_main_architecture}, each of them will be horizontally partitioned into $P$ parts, followed by Horizontal Pooling (HP)~\cite{chao2019gaitset}, Separate Fully Connected layers~\cite{chao2019gaitset}, and BNNeck~\cite{chao2019gaitset} as Gait Head for final feature learning.
        Notably, \textit{the spatial global pooling of each part in HP converts equivariant features into the vectors, supporting our final invariance learning.}

        \noindent\textbf{Loss Function.}
        The loss function follows the current popular gait method~\cite{fan2025opengait} and uses triplet and cross-entropy loss.

\section{Experiments} \label{sec:experiments}
    \subsection{Datasets and Implementation Details}
        \noindent\textbf{Datasets}. We select four of the most recent and challenging gait datasets for evaluation, involving both outdoor and indoor settings, \ie, Gait3D~\cite{zheng2022gait3d}, GREW~\cite{zhu2021GREW}, CCPG~\cite{li2023ccpg}, and SUSTech1K~\cite{shen2023lidargait}. The details are shown in Table~\ref{tab:ImplementationDetails}. We strictly adhere to the established protocols of these datasets, and all data pertaining to human subjects comply with ethical and privacy review requirements.

    \noindent\textbf{Implementation Details.}
        \textbf{{{(1)}}} {{During data pretreatment}}, all silhouettes are normalized to 64$\times$44 resolution.
        \textbf{{{(2)}}} {{During training}}, we strictly follow the official dataset settings~\cite{zheng2022gait3d,zhu2021GREW,li2023ccpg,shen2023lidargait} and popular training strategies from recent gait project (OpenGait~\cite{fan2023opengait,fan2025opengait}).
        These include a ($P, K$) sampling strategy in a mini-batch (sampling $P$ identities with $K$ gait sequences per identity, and each sequence consists of a fixed 30 frames), as well as the optimizer, milestones, and total steps. 
        \textbf{{{(3)}}} {{During test}}, all frames in a gait sequence are sent to the model for evaluation with Euclidean distance similarity.
        \textbf{{{(4)}}} {{For model design}}, 
        the number of layers [$l_1, l_2, l_3, l_4$] in $\mathcal{RRS}$-Gait-S are set to [$1,1,1,1$] for CCPG and SUSTech1K as they are collected from controlled environments, while in $\mathcal{RRS}$-Gait-L are set to [$1,4,4,1$] for Gait3D and [$2,4,4,2$] for GREW to suit uncontrolled collections and large-scale datasets. 
        The ${\rm ReflectConv}$ in Table~\ref{tab:model_main_architecture} is set to 2D-Conv-based~\cite{fan2023opengait} in CCPG and SUSTech1K, and P3D-Conv-based~\cite{fan2025opengait} in Gait3D and GREW.
        Our baseline also follows these settings for a fair comparison, named Baseline-S and Baseline-L.
        The part number in Horizontal Pooling of each feature is set to $P$$=$$16$. The angle limit in Eq. (\ref{eq:theta_limit}) is set as $\theta_{limit}$$=$$40^\circ$. 
        The channel reduction $r$ in Eq. (\ref{eq:CCIM_1},\ref{eq:CCIM_2}) is $4$.
        The weight $\beta$ in the loss function is set to 1.0. The margin in triplet loss is set to 0.2.
        \textbf{(5)} For other details, 
        the parameters and FLOPs are based on the Backbone for consistency across different datasets, as they have varying numbers of classifications that can lead to different results. 
        All results are based on NVIDIA RTX 3090 with PyTorch 2.2.

    \subsection{Comparisons with State-of-the-Art}
        \noindent\textbf{Comparisons in CCPG and SUSTech1K.}
        As show in Table~\ref{tab:CCPG_Results} and Table~\ref{tab:SUSTech1K_results}, $\mathcal{RRS}$-Gait achieves consistent improvement.
        Notably, the data collection settings for CCPG include more diverse scale transformations due to clothing changes and variations in body rotation (left or right) in surveillance camera views.
        SUSTech1K incorporates different camera distances, multiple views, and other various settings.
        Despite these challenges, $\mathcal{RRS}$-Gait achieves substantial performance gains, with an average rank-1 improvement of +6.1\% on CCPG and +4.3\% on SUSTech1K. 

        \noindent\textbf{Comparisons in Gait3D and GREW.} 
        As shown in Table~\ref{tab:Gait3D_GREW_Results}, $\mathcal{RRS}$-Gait achieves superior performance compared to existing gait models on the challenging Gait3D and GREW.
        The performance gains stem from the selected three transformations, which frequently appear in real-world scenarios and often mislead model predictions. Our $\mathcal{R}$eflect-$\mathcal{R}$otate-$\mathcal{S}$cale invariance learning strategy effectively captures these variances and provides stable predictions.
        Notably, the final results achieve a promising 76.7\% and 81.0\% rank-1 on Gait3D and GREW.
        Moreover, $\mathcal{RRS}$-Gait demonstrates higher mAP performance compared to rank-1 (+3.8\% \vs +2.3\% in Gait3D), underscoring its better stability, with a more promising improvement in mAP than rank-1 on Gait3D (+4.4\% \vs +3.3\%).

    \subsection{Ablation Study}
        %
        %
        %
        If not specific, we utilize CCPG and Gait3D to conduct ablation studies. The CCPG results are based on the gait protocol, excluding the identical-views cases. The ${\rm Reflect}$, ${\rm Rotate}$, and ${\rm Scale}$ are short for our designed Reflect Invariance Learning (ReEL), Adaptive Rotate Invariance Learning (RoEL), and Multi-Scale Invariance Learning (SEL) modules.
    
        \noindent\textbf{Ablation of ReEL, RoEL, SEL.} 
        As in Table~\ref{tab:Ablation_Study}, we can observe that the ReEL, RoEL, and SEL modules can progressively improve performance, 
        \ie, they achieve +4.5\%, +0.4\%, +1.2\% rank-1 accuracy in the challenging CL condition in CCPG.
        Meanwhile, the RoEL and SEL modules also achieve +2.6\% and +1.8\% mAP in Gait3D.

        \begin{table}[t]
            \centering
            \caption{Ablations of $\mathcal{RRS}$-Gait. The \textbf{Bold} indicates best result.}
            \label{tab:Ablation_Study}
            \setlength{\tabcolsep}{1.5mm}
            \resizebox{1.0\linewidth}{!}{
                \begin{tabular}{ccccccc}
                \toprule
                \multicolumn{1}{c|}{} & \multicolumn{4}{c|}{CCPG (Gait)} & \multicolumn{2}{c}{Gait3D} \\ \cmidrule{2-7} 
                \multicolumn{1}{c|}{\multirow{-2}{*}{Conditions}} & CL & UP & DN & \multicolumn{1}{c|}{BG} & R-1 & mAP \\ 
                \midrule
                \multicolumn{1}{c|}{{\textbf{$\mathcal{RRS}$-Gait (Ours)}}} & \textbf{78.4} & \textbf{81.8} & \textbf{81.9} & \multicolumn{1}{c|}{\textbf{85.7}} & \textbf{76.7} & \textbf{69.6} \\ 
                \midrule
                \multicolumn{7}{l}{\cellcolor[HTML]{EFEFEF}{1. Ablation of each component}} \\ 
                \midrule
                \multicolumn{1}{l|}{\emph{w/o} SEL} & 77.2 & 81.3 & 81.6 & \multicolumn{1}{c|}{84.6} & 75.9 & 67.8 \\
                \multicolumn{1}{l|}{\emph{w/o} SEL, RoEL} & 76.8 & 80.2 & 81.3 & \multicolumn{1}{c|}{82.8} & - & - \\
                \multicolumn{1}{l|}{\emph{w/o} SEL, RoEL, ReEL} & 72.3 & 76.1 & 76.9 & \multicolumn{1}{c|}{78.9} & 73.4 & 65.2 \\ 
                \midrule
                \multicolumn{7}{l}{\cellcolor[HTML]{EFEFEF}{2. Ablation of angle limit ($\theta_{limit}$) in RoEL}} \\ 
                \midrule
                \multicolumn{1}{l|}{w/ $\theta_{limit}=50^\circ$} & \multicolumn{1}{l}{\textbf{77.3}} & \multicolumn{1}{l}{\textbf{81.3}} & \multicolumn{1}{l}{81.5} & \multicolumn{1}{l|}{84.3} & 75.8 & 67.6 \\
                \multicolumn{1}{l|}{\textbf{w/ $\theta_{limit}=40^\circ$}} & 77.2 & \textbf{81.3} & \textbf{81.6} & \multicolumn{1}{c|}{\textbf{84.6}} & \textbf{75.9} & \textbf{67.8} \\
                \multicolumn{1}{l|}{w/ $\theta_{limit}=30^\circ$} & 75.6 & 78.3 & 78.1 & \multicolumn{1}{c|}{82.2} & 75.4 & 67.1 \\
                \multicolumn{1}{l|}{w/ $\theta_{limit}=20^\circ$} & 75.6 & 78.4 & 78.8 & \multicolumn{1}{c|}{81.5} & 74.8 & 66.3 \\ 
                \midrule
                \multicolumn{7}{l}{\cellcolor[HTML]{EFEFEF}{3. Ablation of each component in SEL}} \\ 
                \midrule
                \multicolumn{1}{l|}{\textbf{\emph{w/} Stage 1-2-3-4}} & \textbf{78.4} & \textbf{81.8} & \textbf{81.9} & \multicolumn{1}{c|}{\textbf{85.7}} & \textbf{76.7} & \textbf{69.6} \\
                \multicolumn{1}{l|}{\emph{w/} Stage 2-3-4} & 77.1 & 81.2 & 80.7 & \multicolumn{1}{c|}{85.4} & 76.1 & 68.5 \\ 
                \midrule
                \multicolumn{1}{l|}{\emph{w/o} Cross-Channel Interaction} & 77.8 & 81.4 & 81.7 & \multicolumn{1}{c|}{85.6} & 76.0 & 69.2 \\ 
                \multicolumn{1}{l|}{\emph{w/o} Cross-Scale Interaction} & 77.6 & 81.5 & 81.4 & \multicolumn{1}{c|}{85.4} & 74.6 & 67.1 \\ 
                \multicolumn{1}{l|}{\emph{w/o} Gate Attention} & 77.8 & 81.7 & 81.8 & \multicolumn{1}{c|}{85.2} & 76.2 & 69.2 \\ 
                \bottomrule
                \end{tabular}
            }
        \end{table}

        \begin{table}[]
            \centering
            \setlength{\tabcolsep}{1.7mm}
            \caption{Ablation of GPool in ${\rm Reflect}$ Module. C=32$_{\times 2}$ means 32 channels are from regular kernels and another 32 are from reflected kernels. ${\rm GPool}$ aggregates regular and reflected features along channels before feeding them into the fully connected layer. The \textbf{Bold} indicates best result.}
            \label{tab:Ablation_GPool}
            \resizebox{1.0\linewidth}{!}{
                \begin{tabular}{l|cc|cccc}
                \toprule
                \multicolumn{1}{c|}{\multirow{2}{*}{Condition}} & \multicolumn{2}{c|}{Backbone} & \multicolumn{4}{c}{CCPG (R-1)} \\ \cmidrule{2-7} 
                \multicolumn{1}{c|}{} & \multicolumn{1}{c|}{\#Last Channels} & \#FC Layer & CL & UP & DN & BG \\ 
                \midrule
                Baseline & \multicolumn{1}{c|}{C=512} & 512$\rightarrow$256 & 72.3 & 76.1 & 76.9 & 78.9 \\
                \midrule
                \emph{w/o} GPool & \multicolumn{1}{c|}{C=256$_{\times 2} =$ 512} & 512$\rightarrow$256 & 75.8 & 78.5 & 79.7 & \textbf{83.4} \\
                \rowcolor[HTML]{ECF4FF} 
                \textbf{\emph{w/} GPool} & \multicolumn{1}{c|}{\textbf{C=256$_{\times 2}$ $\rightarrow$ 256}} & \textbf{256$\rightarrow$256} & \textbf{76.8} & \textbf{80.2} & \textbf{81.3} & 82.8 \\ 
                \bottomrule
                \end{tabular}
            }
        \end{table}
        \noindent\textbf{Ablation of Group Pool in ReEL module.}
        GPool is crucial to ensure reflect equivariance. 
        Although its necessity is theoretically guaranteed, we also conduct the ablation study on GPool. 
        Table~\ref{tab:Ablation_GPool} shows that model performance drops by -1.0\% (CCPG-CL) without GPool but remains +3.5\% higher than the baseline. 
        This confirms the theoretical necessity and practical importance of GPool in the ReEL module.

        \noindent\textbf{Ablation of rotate limit ($\theta_{limit}$) in RoEL module.}
        Due to the distinctive upright walking pattern of humans,  the deviation angles to the left or right are always small. 
        Even under abnormal surveillance angles, the deviation remains limited (\ie, the angle between human body and vertical axis in an image is limited).
        We define this angle limit as a hyper-parameter and ablation results in Table~\ref{tab:Ablation_Study} show that 40$^{\circ}$ limit is effective, with larger values (\eg $\geq50^{\circ}$) providing no additional benefit as performance stabilizes.

        \noindent\textbf{Ablation of SEL module.}
        To achieve effective multi-scale interaction, we design three sub-modules: 
        Selecting features from various layers with different spatial scales,  
        cross-channel and cross-scale interactions to adapt to different scales, 
        and gate attention to maintain training stability.
        Ablation results in Table~\ref{tab:Ablation_Study} show that: 
        (1) Select layers 1-2-3-4 is more effective than layers 2-3-4 (69.6\% \vs 68.5\% mAP on Gait3D); 
        (2) Both cross-channel and cross-scale interactions provide better performances, with cross-scale interactions yielding greater gains;
        (3) Gate attention also proves effective on both datasets.
        
\begin{figure}[t]
        \centering
        \includegraphics[width=1.0\linewidth]{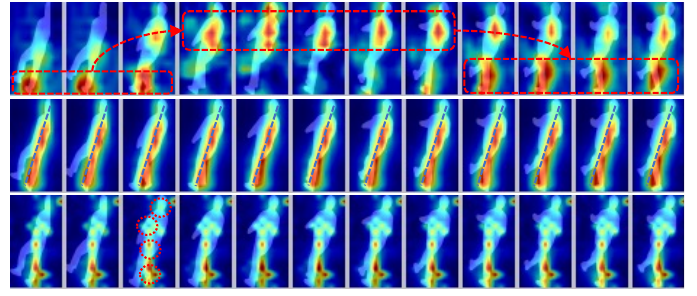}
        \caption{Visualizations of ReEL, RoEL, and SEL outputs (from up to bottom). RoEL or SEL shares the same heatmap across frames since their features undergo temporal pooling.}
        \label{fig:visualization}
    \end{figure}

        \noindent\textbf{Visualization of each module.}
        Fig.~\ref{fig:visualization} shows that: 
        (1) The ReEL module can flexibly adapt to each frame and achieve a periodic-like response by focusing on specific actions at particular moments.
        (2) The output of RoEL module shows a clear lean-right response due to the application of rotated convolution with the predicted angle.
        (3) The SEL module captures more fine-grained and key-region responses to form a more detailed representation.

    \begin{table}[]
        \centering
        \setlength{\tabcolsep}{1.5mm}
        \caption{Comparisons with Backbone Parameters (M) and FLOPs (G). The results are based on \underline{30 frames} with \underline{64$\times$44 resolution}. The \textbf{Bold} indicates the best result.}
        \label{tab:Paras_FLOPs}
        \resizebox{1.0\linewidth}{!}{
            \begin{tabular}{l|cc|cccc}
            \toprule
            \multicolumn{1}{c|}{\multirow{2}{*}{Condition}} & \multicolumn{2}{c|}{Backbone} & \multicolumn{4}{c}{CCPG (R-1)} \\ \cmidrule{2-7} 
            \multicolumn{1}{c|}{} & \#Paras. & \#FLOPs & CL & UP & DN & BG \\ 
            \midrule
            Baseline & 4.90M & 35.490G & 72.3 & 76.1 & 76.9 & 78.9 \\ 
            \midrule
            \emph{w/} ReIL & 1.66M & 35.490G & 76.8 & 80.2 & 81.3 & 82.8 \\
            \emph{w/} ReIL, RoIL & 2.25M & 35.491G & 77.2 & 81.3 & 81.6 & 84.6 \\
            \rowcolor[HTML]{ECF4FF} 
            \textbf{\emph{w/} ReIL, RoIL, SIL} & 3.56M & 36.074G & \textbf{78.4} & \textbf{81.8} & \textbf{81.9} & \textbf{85.7} \\ 
            \bottomrule
            \end{tabular}
        }
    \end{table}

    \begin{table}[t]
        \setlength{\tabcolsep}{1.5mm}
        \caption{Cross-domain evaluations under Rank-1 accuracy. The \textbf{Bold} indicates the best result.}
        \label{tab:cross-domain-evaluation}
        \resizebox{1.0\linewidth}{!}{
            \begin{tabular}{c|c|cccc}
            \toprule
            \multirow{2}{*}{\begin{tabular}[c]{@{}c@{}}\\Source\\ Dataset\end{tabular}} & \multirow{2}{*}{\begin{tabular}[c]{@{}c@{}}\ \\ Method\end{tabular}} & \multicolumn{4}{c}{Target Dataset} \\ \cmidrule{3-6} 
             &  & \multicolumn{1}{c|}{\begin{tabular}[c]{@{}c@{}}CCPG\\ (Gait-CL)\end{tabular}} & \multicolumn{1}{c|}{\begin{tabular}[c]{@{}c@{}}SUSTech1K\\ (Overall)\end{tabular}} & 
             \multicolumn{1}{c|}{\begin{tabular}[c]{@{}c@{}}Gait3D\\ (Rank-1)\end{tabular}}
             & \begin{tabular}[c]{@{}c@{}}GREW\\ (Rank-1)\end{tabular} \\ 
             \midrule
             \multirow{2}{*}{CCPG} & Baseline & \multicolumn{1}{c|}{72.3} & \multicolumn{1}{c|}{17.4} & \multicolumn{1}{c|}{13.0} & 13.2 \\
             & $\mathcal{RRS}$-Gait & \multicolumn{1}{c|}{\textbf{78.4}} & \multicolumn{1}{c|}{\textbf{26.7}} & \multicolumn{1}{c|}{\textbf{19.2}} & \textbf{16.7} \\ 
             \midrule
             \multirow{2}{*}{SUSTech1K} & Baseline & \multicolumn{1}{c|}{16.3} & \multicolumn{1}{c|}{76.0} & \multicolumn{1}{c|}{12.9} & 12.2 \\
             & $\mathcal{RRS}$-Gait & \multicolumn{1}{c|}{\textbf{20.4}} & \multicolumn{1}{c|}{\textbf{80.3}} & \multicolumn{1}{c|}{\textbf{20.9}} & \textbf{15.7} \\ 
             \midrule
            \multirow{2}{*}{Gait3D} & Baseline & \multicolumn{1}{c|}{13.4} & \multicolumn{1}{c|}{50.2} & \multicolumn{1}{c|}{73.4} & 35.5 \\
             & $\mathcal{RRS}$-Gait & \multicolumn{1}{c|}{\textbf{15.5}} & \multicolumn{1}{c|}{\textbf{50.5}} & \multicolumn{1}{c|}{\textbf{76.7}} & \textbf{36.9} \\ 
             \midrule
            \multirow{2}{*}{GREW}  & Baseline & \multicolumn{1}{c|}{12.2} & \multicolumn{1}{c|}{29.1} & \multicolumn{1}{c|}{32.6} & 79.2 \\
             & $\mathcal{RRS}$-Gait & \multicolumn{1}{c|}{\textbf{13.3}} & \multicolumn{1}{c|}{\textbf{29.4}} & \multicolumn{1}{c|}{\textbf{34.6}} & \textbf{81.0} \\ 
             \bottomrule
            \end{tabular}
        }
    \end{table}

    \begin{figure}[t]
        \centering
        \includegraphics[width=1.0\linewidth]{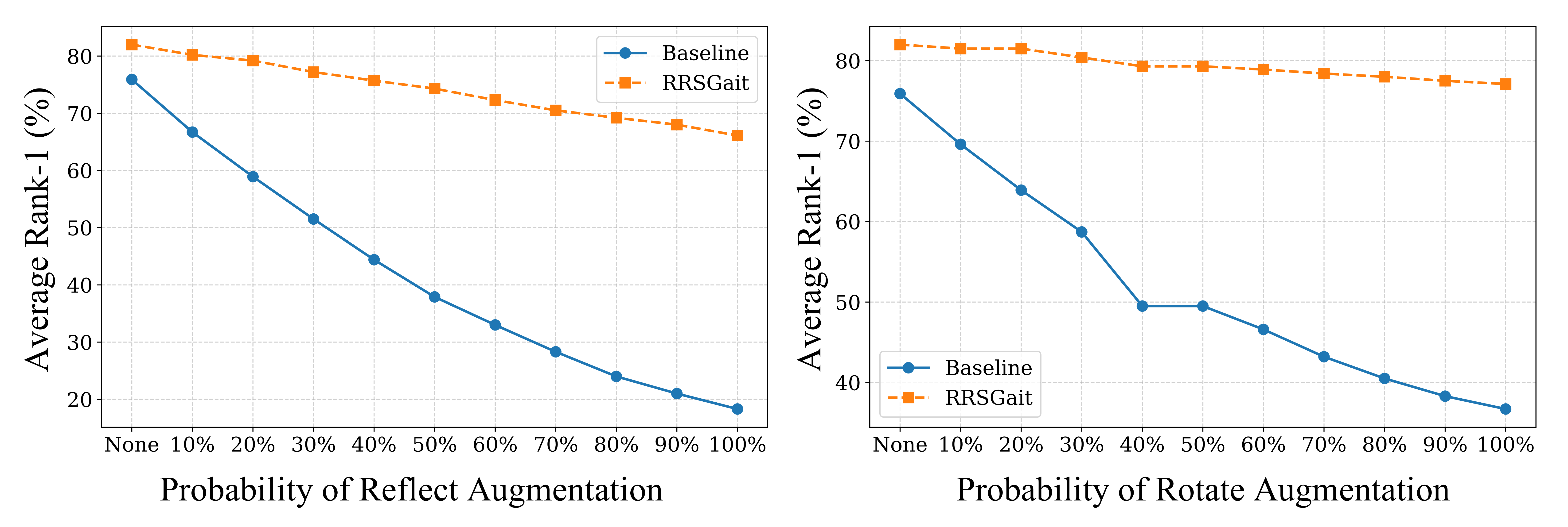}
        \caption{Geometric invariance evaluations through applying varying probabilities of data augmentation (\ie, random reflection and rotation) during the test stage on CCPG.}
        \label{fig:augmentation}
    \end{figure}

    \subsection{Discussion}
        \noindent\textbf{Discussion about Computational Cost}
        As stated in the manuscript, we halve the number of kernels in ReIL module to maintain efficiency.
        Table~\ref{tab:Paras_FLOPs} shows that this modification can maintain the FLOPs (35.490G remains 35.490G), reduce parameters (4.90M$\rightarrow$1.66M), but still achieve high performance (72.3\%$\rightarrow$76.8\%, CCPG-CL).
        Additionally, the input features undergo Temporal Max Pooling to reduce the temporal dimension in RoIL and SIL modules. Even with an increase in parameters (0.59M and 1.31M), computational efficiency is maintained, with only a minimal increase in FLOPs (0.001G and 0.583G).
        
        \noindent\textbf{More Results about Cross-domain Evaluation}
        As shown in Table~\ref{tab:cross-domain-evaluation}, compared to our baseline, $\mathcal{RRS}$-Gait also achieves better generalization results across all four gait datasets. These results further validate the applicability of our model under a broader range of gait conditions.

        \noindent\textbf{More results about geometric invariance.}
        As in Fig.~\ref{fig:augmentation}, to verify our $\mathcal{RRS}$-Gait can effectively mitigate the geometric transformations, we introduce varying levels (\ie, probabilities) of data augmentation (\ie, random reflection or rotation) during the test stage. The results show that $\mathcal{RRS}$-Gait remains stable, with only a slight drop in performance.
        %
        We omit scale augmentation due to the current gait field lacking a standard scale augmentation, leaving it for further exploration.
        
        \noindent\textbf{Limitations.}
        (1) As an initial attempt to leverage geometric invariance to achieve identity invariance, we select three common geometric transformations. More geometric transformations are worth further research to better develop this approach.
        (2) We design distinct equivariant kernels to suit the proposed geometric transformations. It is worthwhile to explore a universal kernel design or aggregate these equivariant kernels into a compact kernel like structural re-parameterizing~\cite{ding2022Scaling}. 

\section{Conclusion} \label{sec:conclusion}
    In this paper, we propose a new perspective: different gait conditions can be approximately linked through geometric transformations, thereby enabling identity invariance in gait recognition via geometric invariance.
    As an initial attempt, we explore three geometric transformations (\ie, reflect, rotate, and scale), and propose a $\mathcal{R}$eflect-$\mathcal{R}$otate-$\mathcal{S}$cale invariance learning framework, named $\mathcal{RRS}$\textbf{-Gait}.
    Extensive experiments and visualizations show consistent improvements in both indoor and outdoor gait datasets.
    We hope this new perspective will inspire further exploration of more geometric transformations to connect different gait conditions for better gait recognition.




%

{
\bibliographystyle{IEEEtran}
\bibliography{RRSGait}
}












\end{document}